\documentclass[letterpaper]{article} % DO NOT CHANGE THIS
\usepackage{aaai2026}  % DO NOT CHANGE THIS
\usepackage{times}  % DO NOT CHANGE THIS
\usepackage{helvet}  % DO NOT CHANGE THIS
\usepackage{courier}  % DO NOT CHANGE THIS
\usepackage[hyphens]{url}  % DO NOT CHANGE THIS
\usepackage{graphicx} % DO NOT CHANGE THIS
\usepackage{natbib} 
\usepackage{caption} 
\usepackage{amsmath, amssymb}
\usepackage{array}
\usepackage[hidelinks]{hyperref}
\usepackage{multirow}
\usepackage{booktabs}
\newcommand{\arbiviewgen}{ArbiViewGen}
\usepackage{xcolor}
\usepackage{amsmath}

\usepackage{algorithm}
\usepackage{algorithmic}
\usepackage{newfloat}
\usepackage{listings}
\DeclareCaptionStyle{ruled}{labelfont=normalfont,labelsep=colon,strut=off} % DO NOT CHANGE THIS
\floatstyle{ruled}
\newfloat{listing}{tb}{lst}{}
\floatname{listing}{Listing}
\title{ArbiViewGen: Controllable Arbitrary Viewpoint Camera Data Generation for Autonomous Driving via Stable Diffusion Models}

\author {
    Yatong Lan\textsuperscript{\rm 1,\rm 2,\rm 3},
    Jingfeng Chen\textsuperscript{\rm 1,\rm 2,\rm 4},
    Yiru Wang\textsuperscript{\rm 2,\rm 5},
    Lei He\textsuperscript{\rm 1,\rm 2}\thanks{Corresponding author: helei2023@tsinghua.edu.cn}
}

\affiliations {
    \textsuperscript{\rm 1}School of Vehicle and Mobility, Tsinghua University, Beijing 100084, China\\
    \textsuperscript{\rm 2}State Key Laboratory of Intelligent Green Vehicle and Mobility, Tsinghua University, Beijing 100084, China\\
    \textsuperscript{\rm 3}School of Science, Minzu University, Beijing 100081, China\\
    \textsuperscript{\rm 4}School of Computer Science, Carnegie Mellon University, Pittsburgh, 15289, USA \\
    \textsuperscript{\rm 5}The Hong Kong University of Science and Technology \\
}

\begin{document}

\maketitle

\begin{abstract}
Arbitrary viewpoint image generation holds significant potential for autonomous driving, yet remains a challenging task due to the lack of ground-truth data for extrapolated views, which hampers the training of high-fidelity generative models. In this work, we propose \arbiviewgen\, a novel diffusion-based framework for the generation of controllable camera images from arbitrary points of view. To address the absence of ground-truth data in unseen views, we introduce two key components: Feature-Aware Adaptive View Stitching (FAVS) and Cross-View Consistency Self-Supervised Learning (CVC-SSL). FAVS employs a hierarchical matching strategy that first establishes coarse geometric correspondences using camera poses, then performs fine-grained alignment through improved feature matching algorithms, and identifies high-confidence matching regions via clustering analysis. Building upon this, CVC-SSL adopts a self-supervised training paradigm where the model reconstructs the original camera views from the synthesized stitched images using a diffusion model, enforcing cross-view consistency without requiring supervision from extrapolated data. Our framework requires only multi-camera images and their associated poses for training, eliminating the need for additional sensors or depth maps. To our knowledge, \arbiviewgen\ is the first method capable of controllable arbitrary view camera image generation in multiple vehicle configurations.
\end{abstract}

\section{Introduction}

The automotive industry has witnessed the emergence of end-to-end autonomous driving technology as a predominant development direction. However, the heterogeneous configurations of multi-source sensor systems have introduced coupling challenges. Models trained with different sensor combinations are difficult to transfer and reuse across platforms. Current autonomous driving systems typically employ multi-camera surround-view configurations as the core perception module, but there are significant differences among vehicle types in terms of the number of cameras, installation positions, and fields of view. These configuration discrepancies result in severely compromised cross-platform data reusability, necessitating extensive data collection and annotation efforts for each new vehicle model, which leads to high development costs and long cycles. 

To address this issue, arbitrary view camera image generation technology has emerged. By generating high-quality images from arbitrary poses using a limited set of existing camera views, it is possible to achieve data reuse across different vehicle types and reduce the development cost for new models. However, unlike general scene reconstruction, data collection in autonomous driving scenarios is usually limited to a single driving trajectory, resulting in sparsity and homogeneity of observed data in 3D space. This is particularly problematic for novel view synthesis, where there is a severe lack of ground truth in extrapolated views: when the rendered viewpoint deviates from the recorded trajectory, it is impossible to obtain ground truth images for direct supervised training, which has become a core bottleneck restricting the development of this technology.

Despite rapid progress, existing multiview image generation methods remain fundamentally limited by their reliance on ground-truth supervision at target viewpoints, a resource that is inherently scarce in autonomous driving scenarios. Current approaches can be roughly grouped into two categories: diffusion-based generation methods and 3D reconstruction-based synthesis methods. Diffusion-based methods (e.g., MVDiffusion~\cite{Tang2023}, SyncDreamer~\cite{Liu2023a}, FreeVS~\cite{Wang2024}, DiST-4D~\cite{Guo2025}) typically adopt an end-to-end paradigm that learns mappings between input-output view pairs, but their performance often suffers in scenarios with sparse or incomplete viewpoint coverage, which is common in driving datasets. 3D reconstruction-based methods (e.g., 3D Gaussian Splatting~\cite{Kerbl2023}, NeRF~\cite{Mildenhall2020}) utilize explicit or implicit scene geometry to enable novel view synthesis, yet their two-stage reconstruction-rendering pipeline is highly sensitive to the spatial sparsity of input views, leading to artifacts and degraded quality in extrapolated regions. These limitations underscore the core challenge: the absence of ground-truth data from novel viewpoints prevents most existing methods from being trained effectively in such scenarios, especially in real-world autonomous driving settings.

We break the dependency on ground-truth supervision at novel viewpoints by introducing CVC-SSL, a self-supervised framework that allows closed-loop training for arbitrary-view generation. Our approach constructs pseudo-novel views via geometric image stitching and employs a diffusion model to reconstruct original camera images from these synthetic views. The reconstruction errors serve as self-supervised signals, allowing the model to learn cross-view geometric relationships and visual consistency without requiring ground-truth data at extrapolated poses. Notably, our method requires only six camera images and their corresponding pose information to achieve end-to-end model training, establishing for the first time a controllable arbitrary viewpoint generation system for multi-vehicle architectures. For extrapolated viewpoints, we propose a quantitative evaluation strategy based on projecting colored LiDAR point clouds to novel views to obtain sparse ground-truth pixels.

The main contributions of this study are as follows.
\begin{itemize}
\item A pure visual image stitching algorithm is developed by combining geometric transformation with hierarchical feature matching. It enables the automatic construction of high-quality pseudo-ground truth data for extrapolated views via precise alignment and texture fusion, offering reliable supervision for training.
\item A self-supervised learning paradigm based on cyclic reconstruction is introduced which establishes bidirectional mappings across views. This design effectively overcomes the lack of ground truth supervision in novel viewpoints and substantially enhances generation quality.
\item To enable quantitative evaluation, a novel image quality assessment strategy is proposed, which projects colored point clouds—sampled from real images—into target views. This establishes the first end-to-end evaluation framework for controllable arbitrary-view generation across diverse vehicle architectures.
\end{itemize}

\begin{figure*}[ht]
\centering
\includegraphics[width=1.0\textwidth]{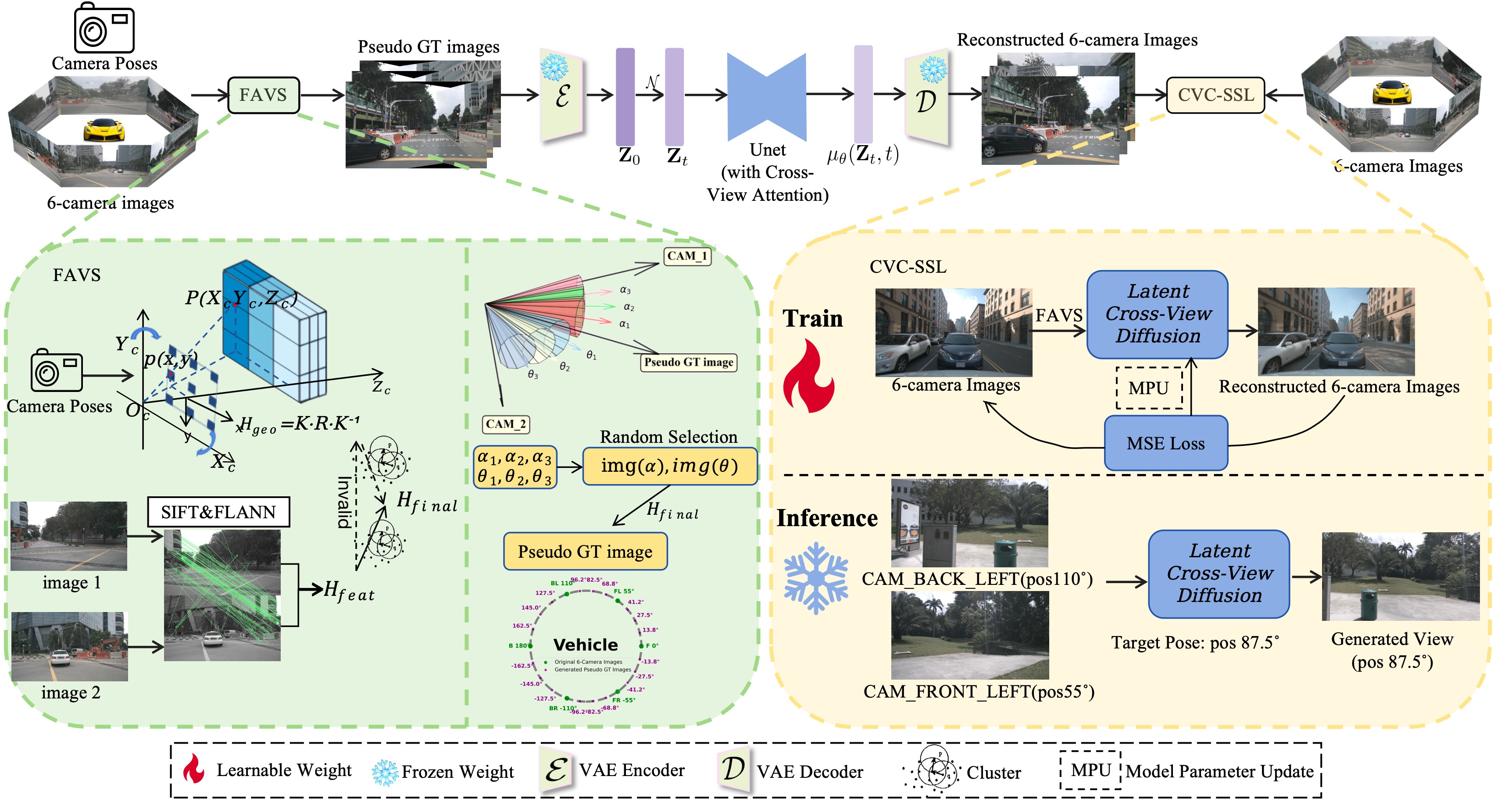}
\caption{\textbf{Pipeline of ArbiViewGen for controllable arbitrary-view image generation.} FAVS generates pseudo ground-truth views via geometry-guided feature stitching. CVC-SSL trains a latent diffusion model with cross-view consistency to generate multi-view images from arbitrary poses using only 6-camera inputs and pose information—without requiring ground-truth extrapolated views.}
\label{fig:1}
\end{figure*}

\section{Related Work}
\subsubsection{Diffusion-based Generation Methods}\quad
Diffusion-based novel view synthesis methods have gained significant attention due to their ability to model complex scene distributions. However, most of these techniques rely heavily on ground-truth supervision at specific training viewpoints, which severely restricts their ability to generalize to unseen or extrapolated views. This limitation becomes particularly evident in autonomous driving scenarios, where such detailed supervision is often unavailable for novel viewpoints. Notable methods include Zero-1-to-3~\cite{Liu2023b}, which integrates camera pose embeddings to enhance the synthesis of new perspectives, though it is primarily designed for object-centric scenes; StreetCrafter~\cite{Yan2024}, which leverages a LiDAR-conditioned video diffusion approach to generate novel views, yet depends heavily on the availability of additional sensors and is less robust when tasked with synthesizing views at large angular extrapolations; DiST-4D~\cite{Guo2025}, which incorporates metric depth information to facilitate 4D scene synthesis, but its reliance on precise depth data limits its applicability in the absence of such data; and DriveX~\cite{Yang2024} and Drive-1-to-3~\cite{Lin2024}, which enhance the synthesis quality within constrained camera setups, yet still fail to guarantee geometric consistency when applied to novel viewpoints lacking supervision. The underlying challenge these methods face is their dependence on explicit ground-truth supervision, which fundamentally constrains their ability to perform well in situations where data is sparse or entirely missing from new viewpoints. In contrast, our framework addresses this limitation by adopting a self-supervised learning paradigm, which allows the generation of arbitrary viewpoints without requiring any ground-truth supervision, thereby expanding the potential for real-world applications in autonomous driving.

\subsubsection{3D Reconstruction-based Novel View Synthesis}\quad
In contrast to diffusion-based methods, 3D reconstruction-based approaches, such as Neural Radiance Fields (NeRF)\cite{Mildenhall2020} and 3D Gaussian Splatting (3DGS)\cite{Kerbl2023}, harness geometric priors and volumetric scene representations to facilitate novel view synthesis. These methods have shown great promise in static environments where dense, overlapping observations are available to estimate geometry. However, their performance degrades significantly in dynamic, large-scale urban environments, where sparse observations and the complexity of real-world scenes pose significant challenges. Recent innovations like S³Gaussian~\cite{Huang2024}, SplatFlow~\cite{Sun2024}, and EVolSplat~\cite{Miao2025} have made strides in improving reconstruction fidelity, but they still struggle with large-angle extrapolation due to spatial sparsity and a lack of sufficient constraints. This often results in visible artifacts and reduced quality in synthesized views. To mitigate these issues, methods like VEGS~\cite{Hwang2024} and DHGS~\cite{Shi2024} integrate LiDAR data or adopt hybrid fusion strategies that combine multiple sensor modalities, enhancing the robustness of the synthesis process. However, these approaches remain heavily reliant on external sensors or high-quality point clouds, making them less adaptable in environments where such data is not available or is incomplete. In contrast, our method sidesteps these challenges by directly learning to synthesize arbitrary views from scene data in a fully self-supervised manner, eliminating the need for auxiliary data and ensuring broader applicability across various environments, including those with limited sensor input or sparse observations.
\section{Methods}
We introduce the design of our proposed \arbiviewgen\ in this section, where the overall pipeline is in Figure~\ref{fig:1}.

\subsection{Overview}
\arbiviewgen\ addresses the challenging problem of arbitrary viewpoint image generation in multi-vehicle autonomous driving scenarios, where the lack of ground truth for extrapolated viewpoints poses significant difficulties. Our approach can synthesis high-quality camera images from arbitrary target viewpoints based on limited camera viewpoints. The proposed method consists of two core modules: \textbf{Feature-Aware Adaptive View Synthesis (FAVS)} and \textbf{Cross-View Consistency Self-Supervised Learning (CVC-SSL)}. The FAVS module uses only visual inputs to generate high-quality pseudo ground truth for novel viewpoints by combining geometric constraints and multi-scale feature-level cues.  
The CVC-SSL module is built on latent diffusion models with a cross-view consistency attention mechanism. The pseudo ground truth is constructed from real images and used as input, while the real images themselves serve as supervision, forming a self-supervised training loop. Leveraging the generative capability of the attention-based model, our framework effectively extrapolates high-quality novel viewpoints by referencing limited number of real images.

\subsection{Feature-Aware Adaptive View Synthesis (FAVS) Algorithm}
The core idea of the FAVS algorithm is to achieve high-quality stitching of six camera images to arbitrary target viewpoints through a hierarchical optimization strategy. The algorithm consists of four progressive optimization stages: geometric transformation establishment, feature matching optimization, object alignment fine-tuning, and adaptive fusion generation. This hierarchical design ensures progressive optimization from coarse to fine, effectively addressing geometric consistency and visual quality issues in complex driving scenarios.

\subsubsection{Stage 1: Geometric Transformation Foundation}

Following principles of camera geometry, we establish the mathematical mapping between different viewpoints using homography. Given the intrinsic matrix $\mathbf{K}_1$ and rotation matrix $\mathbf{R}_1$ of the source camera, and the intrinsic matrix $\mathbf{K}_2$ and rotation matrix $\mathbf{R}_2$ of the target camera, the transformation between views under a pure rotation assumption is formulated as:
\begin{equation*}
\mathbf{H}_{\text{geom}} = \mathbf{K}_2 \mathbf{R}_2 \mathbf{R}_1^{-1} \mathbf{K}_1^{-1}
\end{equation*}
Here, the camera intrinsic matrix $\mathbf{K}$ is defined by the focal length $f$ and principal point $(c_x, c_y)$:
\begin{equation*}
\mathbf{K} = 
\begin{bmatrix}
f & 0 & c_x \\
0 & f & c_y \\
0 & 0 & 1
\end{bmatrix}
\end{equation*}
The camera rotation matrix $\mathbf{R}$ is parameterized by azimuth angle $\theta$ and elevation angle $\phi$, and is constructed as:
\begin{equation*}
\mathbf{R} = \mathbf{R}_z(\theta)\, \mathbf{R}_x(\phi)
\end{equation*}
where $\mathbf{R}_z(\theta)$ and $\mathbf{R}_x(\phi)$ denote standard rotation matrices around the $z$-axis and $x$-axis, respectively. 

Although the source and target cameras may have a relative translation $\mathbf{t}_{j,i}$, in autonomous driving scenarios, most objects are typically far from the camera. %When the scene depth $d$ satisfies the far-field condition $\frac{d}{\|\mathbf{t}_{j,i}\|} \gg 1$, the effect of translation on image coordinates becomes negligible.%
Under this approximation, the homography matrix can be simplified to a rotation-only form:
\begin{equation*}
\mathbf{H}_{\text{geom}} \approx \mathbf{K}_2 \mathbf{R}_{j,i} \mathbf{K}_1^{-1}, \quad \text{where } \mathbf{R}_{j,i} = \mathbf{R}_2 \mathbf{R}_1^{-1}
\end{equation*}

It is important to distinguish this approximation from the planar scene assumption, where all 3D points are constrained to lie on a single plane. The planar assumption leads to a different homography formulation that explicitly incorporates the plane's normal vector $\mathbf{n}$ and its distance from the camera $d$, typically written as:
\begin{equation*}
\mathbf{H}_{\text{planar}} = \mathbf{K}_2 \left( \mathbf{R} - \frac{\mathbf{t} \mathbf{n}^\top}{d} \right) \mathbf{K}_1^{-1}
\end{equation*}
In contrast, our method relies solely on the far-field approximation, avoiding the need to estimate depth or plane parameters. To ensure geometric consistency, we validate the computed homography by checking the transformation of the four image corner points to confirm that no excessive distortion occurs:
$\text{validity} = \text{check\_homography}(\mathbf{H}_{\text{geom}}, \text{corners})$

\subsubsection{Stage 2: Feature Matching Optimization}

After obtaining the basic geometric transformation, we introduce a SIFT feature-based matching mechanism to optimize transformation parameters. First, SIFT feature extraction is performed on source and reference images:
\begin{equation*}
\{\operatorname{kp}_1, \operatorname{des}_1\} = \operatorname{SIFT}(I_{\operatorname{source}}),  
\{\operatorname{kp}_2, \operatorname{des}_2\} = \operatorname{SIFT}(I_{\operatorname{reference}})
\end{equation*}
FLANN matcher is used to establish feature point correspondences, and high-quality matches are filtered through Lowe's ratio test:
\begin{equation*}
\operatorname{good\_matches} = \left\{ m \;:\; \frac{d_1}{d_2} < 0.75 \right\}
\end{equation*}
where $d_1$ and $d_2$ are the distances to the nearest and second-nearest neighbors, respectively. The precise homography matrix is estimated through RANSAC algorithm:
\begin{equation*}
H_{\operatorname{feature}} = \operatorname{RANSAC}(\operatorname{good\_matches})
\end{equation*}
At this stage, we evaluate whether the feature-based matching result is reliable to refine the transformation. The decision is made based on the following criteria:
\begin{itemize} 
\setlength{\parsep}{0pt}
\raggedright 
    \item Number of matching points: $|\mbox{good\_matches}|\geq\mbox{min\_matches}$
    \item Homography matrix validity:
    $\mbox{check\_homography}(H_{feature})$
    \item Consistency with geometric transformation: $\mbox{consistency}(H_{geometric},H_{feature})<\mbox{threshold}$
\end{itemize}
If feature matching satisfies all conditions, the base transformation is updated:
\begin{equation*}
H_{\operatorname{base}} = 
\begin{cases}
H_{\operatorname{feature}} & \operatorname{if\ consistent} \\
\begin{aligned}
\alpha H_{\operatorname{geometric}} + {} \\
(1 - \alpha) H_{\operatorname{feature}}
\end{aligned} & \operatorname{if\ partially\ consistent} \\
H_{\operatorname{geometric}} & \operatorname{otherwise}
\end{cases}
\end{equation*}
where the weight $\alpha$ is dynamically adjusted based on the consistency degree.

\subsubsection{Stage 3: Object Alignment Fine-tuning}

To ensure precise correspondence of important objects across different viewpoints, we introduce a DBSCAN-based object detection and alignment mechanism:
\begin{equation*}
\text{clusters} = \text{DBSCAN}(\text{keypoints}, \epsilon, \text{min\_samples})
\end{equation*}
where $\epsilon$ is the clustering radius and $\mbox{min\_samples}$ is the minimum number of samples. For each detected object cluster, we compute its features:
%$$\mbox{object}_j = \{\mbox{center}_j, \mbox{bbox}_j, \mbox{confidence}_j, \mbox{type}_j\}$$
\begin{equation*}
\text{object}_j = \{\text{center}_j, \text{bbox}_j, \text{confidence}_j, \text{type}_j\}
\end{equation*}
The object center is calculated as the centroid of all feature points within the cluster. For the $j$-th cluster containing $n_j$ feature points, with feature point coordinates in source and target images denoted as $\{p_{1i}\}_{i=1}^{n_j}$ and $\{p_{2i}\}_{i=1}^{n_j}$ respectively, the object center is computed as:
$c_{1j} = \frac{1}{n_j}\sum_{i=1}^{n_j} p_{1i}, \quad c_{2j} = \frac{1}{n_j}\sum_{i=1}^{n_j} p_{2i}$
where $c_{1j}$ and $c_{2j}$ are the center coordinates of the $j$-th object in source and target images.
Object alignment is achieved by minimizing weighted centroid offset to ensure geometric consistency of key objects:
%$$\Delta T = \arg\min_{\Delta T} \sum_{j=1}^m w_j \|c_{2j} - (H_{base} \cdot c_{1j} + \Delta T)\|^2$$
\begin{equation*}
\Delta T = \arg\min_{\Delta T} \sum_{j=1}^{m} w_j \left\| c_{2j} - \left( H_{\text{base}} \cdot c_{1j} + \Delta T \right) \right\|^2
\end{equation*}
The weight $w_j$ considers object type and feature point density:
$w_j = \mbox{confidence}_j \times \mbox{type\_weight}_j \times \mbox{density\_factor}_j$
The $\mbox{confidence}_j$ is the cluster confidence, $\mbox{type\_weight}_j$ is the object type weight (e.g., vehicles, pedestrians), and $\mbox{density\_factor}_j$ is the feature point density factor.
The final alignment transformation matrix is:
%$$H_{aligned} = \left[\begin{array}{ccc} 1 & 0 & \beta \Delta T_x \\ 0 & 1 & \beta \Delta T_y \\ 0 & 0 & 1 \end{array}\right] \cdot H_{base}$$
\begin{equation*}
H_{\text{aligned}} = 
\begin{bmatrix}
1 & 0 & \beta \Delta T_x \\
0 & 1 & \beta \Delta T_y \\
0 & 0 & 1
\end{bmatrix}
\cdot H_{\text{base}}
\end{equation*}
where $\beta$ is the adjustment strength parameter controlling the influence degree of object alignment, with a range of $[0, 1]$.

\subsubsection{Stage 4: Adaptive Fusion Generation}

Each candidate image is transformed using the corresponding transformation:
%$$I_{warped}^{(i)} = \mbox{warp\_perspective}(I_{source}^{(i)}, H_{final}^{(i)})$$
\begin{equation*}
I_{\text{warped}}^{(i)} = \text{warp\_perspective}\left(I_{\text{source}}^{(i)}, H_{\text{final}}^{(i)}\right)
\end{equation*}
Fusion weights are determined by multiple factors:
%$$w^{(i)}(x,y)=w_{distance}^{(i)}(x,y)\cdot w_{gradient}^{(i)}(x,y) \cdot w_{quality}^{(i)}\cdot w_{primary}^{(i)}$$
\begin{equation*}
w^{(i)}(x, y) = w_{\text{distance}}^{(i)}(x, y) \cdot w_{\text{gradient}}^{(i)}(x, y) \cdot w_{\text{quality}}^{(i)} \cdot w_{\text{primary}}^{(i)}
\end{equation*}
where:
\begin{itemize} 
\setlength{\parsep}{0pt}
\raggedright 
    \item Distance weight: $w_{distance}(x,y)=\left(\frac{d(x,y)}{d_{max}}\right)^{\gamma}$, weight based on distance transform
    \item Gradient weight: $w_{gradient}(x,y)=\frac{1}{1 + \|\nabla I(x,y)\|/\sigma}$, reducing weight in high-gradient regions
    \item Quality weight: $w_{quality}^{(i)}$, global weight based on matching quality
    \item Primary camera weight: $w_{primary}^{(i)}$, weight bonus for primary camera
\end{itemize}
The final fusion result is obtained through weighted averaging:
\begin{equation*}
I_{\text{target}} = \frac{\sum_{i=1}^n w^{(i)} \cdot I_{\text{warped}}^{(i)}}{\sum_{i=1}^n w^{(i)}}
\end{equation*}
\subsection{Cross-View Consistency Self-Supervised Learning Framework (CVC-SSL)}

To address the problem of lacking ground truth for extrapolated viewpoints, we design the CVC-SSL framework, constructing a closed-loop training mechanism. The core innovation of this framework lies in utilizing diffusion models to inversely reconstruct original viewpoints from stitched extrapolated viewpoint images, forming a self-supervised learning closed loop. The overall training loop is in Algorithm~\ref{alg:cvc_ssl}.

\subsubsection{Self-Supervised Training Process}
During training, we perform the following steps for each training sample:
\begin{itemize}
    \item Use six real camera images $\{I_1, I_2, \ldots, I_6\}$ along with their corresponding pose information $\{P_1, P_2, \ldots, P_6\}$.
    \item For each real image $I_i$ and its pose $P_i$, randomly sample pseudo target poses to the left and right of the original camera position (denoted as $P_{\text{p-left}}$ and $P_{\text{p-right}}$). Note that these pseudo target poses are sampled at different spatial positions to simulate novel viewpoints.
    \item Apply the FAVS algorithm to synthesize pseudo images at the sampled poses:
    \[
    I_{\text{p}} \leftarrow \text{FAVS}(\{I_1, \ldots, I_6\}, \{P_1, \ldots, P_6\}, P_{\text{p}})
    \]
    \item For each real image, use the pseudo images from both sides (left and right) as input to the diffusion model. The model, equipped with geometry-guided cross-view attention, learns to reconstruct the original real image as its prediction target.

\end{itemize}

\subsubsection{Loss Function Design}

We design a multi-level loss function to ensure the model maintains geometric consistency and visual quality while learning to generate multi-view images.
\begin{itemize}
    \item The main reconstruction loss adopts the standard denoising diffusion loss:
    \begin{equation*}
    \mathcal{L}_{\text{main}} = \mathbb{E}_{x_0, \epsilon \sim \mathcal{N}(0, I),\, t} \left[ \left\| \epsilon - \epsilon_\theta(x_t, t, f_{\text{pose}}) \right\|^2 \right]
    \end{equation*}
    $x_0$ is the latent representation of the target viewpoint image generated by the FAVS algorithm, $\epsilon$ is the added noise, $t$ is the diffusion time step, and $f_{\text{pose}}$ is the pose condition encoding.

    \item The geometric consistency loss ensures generated images maintain geometric consistency across different viewpoints:
    \begin{equation*}
    \mathcal{L}_{\text{geo}} = \sum_{i,j} \left\| M_{i,j}^{\text{pred}} - M_{i,j}^{\text{target}} \right\|_F
    \end{equation*}
    $M_{i,j}^{pred}$ is the predicted cross-view attention map, $M_{i,j}^{\text{target}}$ is the target attention map computed based on geometric correspondences, and $\|\cdot\|_F$ denotes the Frobenius norm.

    \item The perceptual quality loss adopts VGG feature-based perceptual loss:
    \begin{equation*}
    \mathcal{L}_{\text{perceptual}} = \sum_{l} \lambda_l \left\| \phi_l(I_{\text{pred}}) - \phi_l(I_{\text{target}}) \right\|_2
    \end{equation*}
    $\phi_l$ represents the feature extractor of the $l$-th layer of the VGG network, and $\lambda_l$ is the corresponding weight coefficient.
\end{itemize}
The total loss function is a weighted combination of the above loss terms:
\begin{equation*}
\mathcal{L}_{\text{total}} = \mathcal{L}_{\text{main}} + \alpha\, \mathcal{L}_{\text{geo}} + \beta\, \mathcal{L}_{\text{perceptual}}
\end{equation*}
where $\alpha = 0.1$ and $\beta = 0.01$ are hyperparameters balancing different loss terms, determined through experiment.

\begin{algorithm}
\caption{CVC-SSL: Batch-wise Self-Supervised Training with Multi-Pair Pseudo Views}
\label{alg:cvc_ssl}
\begin{algorithmic}[1]
\REQUIRE Batch of real images $\{I_i\}_{i=1}^B$, their poses $\{P_i\}_{i=1}^B$, FAVS algorithm, diffusion model $\mathcal{M}$

\FOR{each real image $I_i$ in the batch with pose $P_i$}
    \STATE Sample $K$ pseudo target poses on the left: $\{P^{(k)}_{\text{p-left}}\}_{k=1}^K$
    \STATE Sample $K$ pseudo target poses on the right: $\{P^{(k)}_{\text{p-right}}\}_{k=1}^K$
    \FOR{each pseudo pose pair $(P^{(k)}_{\text{p-left}}, P^{(k)}_{\text{p-right}})$}
        \STATE Generate pseudo images via FAVS using full batch context:
        \STATE \quad $I^{(k)}_{\text{p-left}} \leftarrow \text{FAVS}(\{I_j\}, \{P_j\}, P^{(k)}_{\text{p-left}})$
        \STATE \quad $I^{(k)}_{\text{p-right}} \leftarrow \text{FAVS}(\{I_j\}, \{P_j\}, P^{(k)}_{\text{p-right}})$
        \STATE Feed $(I^{(k)}_{\text{p-left}}, I^{(k)}_{\text{p-right}}, P^{(k)}_{\text{p-left}}, P^{(k)}_{\text{p-right}})$ into model $\mathcal{M}$
        \STATE Predict $\hat{I}_i$ for target view $P_i$
        \STATE Compute total loss $\mathcal{L}_{\text{total}}$ with:
        \STATE \quad $\mathcal{L}_{\text{main}}$: denoising loss
        \STATE \quad $\mathcal{L}_{\text{geo}}$: geometric consistency
        \STATE \quad $\mathcal{L}_{\text{perceptual}}$: VGG perceptual loss
        \STATE \quad $\mathcal{L}_{\text{total}} = \mathcal{L}_{\text{main}} + \alpha \mathcal{L}_{\text{geo}} + \beta \mathcal{L}_{\text{perceptual}}$
        \STATE Update model $\mathcal{M}$ using $\mathcal{L}_{\text{total}}$
    \ENDFOR
\ENDFOR
\end{algorithmic}
\end{algorithm}

\subsection{Geometry-Guided Cross-View Attention Mechanism}

To model correspondences between different viewpoints, we design a multi-view attention mechanism based on geometric constraints.
\subsubsection{Cross-View Geometric Correspondences}
Consider two camera viewpoints $i$ and $j$ observing the same planar scene. For any point $\mathbf{X}$ on the plane, its projection relationship in the two cameras can be described by the homography matrix $H_{i,j}$:
\begin{equation*}
\mathbf{p}_j \sim H_{i,j} \, \mathbf{p}_i
\end{equation*}
where $\mathbf{p}_i$ and $\mathbf{p}_j$ are the homogeneous coordinates of point $\mathbf{X}$ in cameras $i$ and $j$.
\subsubsection{Geometry-Guided Feature Alignment}
For the feature $\mathbf{f}_i(\mathbf{p}_i)$ at position $\mathbf{p}_i$ in viewpoint $i$, we compute its corresponding position in viewpoint $j$:
%$$\mathbf{p}_j^{(l)} = \pi(H_{i,j}^{(l)} \mathbf{p}_i)$$
\begin{equation*}
\mathbf{p}_j^{(l)} = \pi\left(H_{i,j}^{(l)} \, \mathbf{p}_i\right)
\end{equation*}
where $\pi(\cdot)$ represents the conversion from homogeneous coordinates to Cartesian coordinates.
\subsubsection{Multi-Level Attention Computation}
Considering the multi-level structure of the scene, we design a hierarchical attention mechanism to effectively handle geometric relationships at different depth levels:
%$$\mbox{Attention}(\mathbf{Q}_i, \mathbf{K}_j, \mathbf{V}_j) = \sum_{l=1}^{L} w_l \cdot \mbox{Attention}^{(l)}(\mathbf{Q}_i, \mathbf{K}_j^{(l)}, \mathbf{V}_j^{(l)})$$
\begin{equation*}
\text{Attention}(\mathbf{Q}_i, \mathbf{K}_j, \mathbf{V}_j) = \sum_{l=1}^{L} w_l \cdot \text{Attention}^{(l)}(\mathbf{Q}_i, \mathbf{K}_j^{(l)}, \mathbf{V}_j^{(l)})
\end{equation*}
where $\mathbf{K}_j^{(l)} = \mathbf{K}_j \odot \mathbf{M}_j^{(l)}$, $\mathbf{M}_j^{(l)}$ is the mask of the $l$-th layer, $w_l$ is the layer weight, satisfying $\sum_{l=1}^{L} w_l = 1$.

\subsubsection{Geometric Design of Positional Encoding}

To enhance geometric perception capability, we design relative pose-based positional encoding, adding geometric information into attention computation:
%$$\mbox{PE}(\mathbf{p}_i, \mathbf{p}_j) = \mbox{concat}(\sin(\mathbf{W}_1 \Delta\mathbf{p}), \cos(\mathbf{W}_2 \Delta\mathbf{p}))$$
\begin{equation*}
\text{PE}(\mathbf{p}_i, \mathbf{p}_j) = \text{concat}\left( \sin(\mathbf{W}_1 \Delta\mathbf{p}),\; \cos(\mathbf{W}_2 \Delta\mathbf{p}) \right)
\end{equation*}
where $\Delta\mathbf{p} = \mathbf{p}_j - H_{i,j}\mathbf{p}_i$ represents the deviation between geometrically predicted position and actual position.

\subsection{Implementation Details}

\subsubsection{Network Architecture}

We base our architecture on Stable Diffusion's UNet, inserting cross-view attention modules at each level of the encoder, intermediate layers, and decoder. The attention module dimensions are set as: encoder layers (320, 640, 1280, 1280), intermediate layers (1280), decoder layers (1280, 1280, 640, 320). This design ensures effective modeling of cross-view geometric relationships at different resolution levels.

\subsubsection{Training Parameters}

We use the following training parameters: learning rate $1 \times 10^{-4}$ (cosine annealing schedule), batch size 8 (per GPU), diffusion steps 1000 (training) / 50 (inference), guidance strength 7.5. These parameters have been thoroughly validated through experiments, achieving a good balance between generation quality and training efficiency.

Through this design, \arbiviewgen\ can generate high-quality arbitrary viewpoint images using only six camera images and their pose information, effectively solving the problem of lacking ground truth for extrapolated viewpoints, providing a feasible technical solution for multi-vehicle data reuse in autonomous driving scenarios.

\section{Experiment}
\begin{figure*}[t]
\centering
\includegraphics[width=1.0\textwidth]{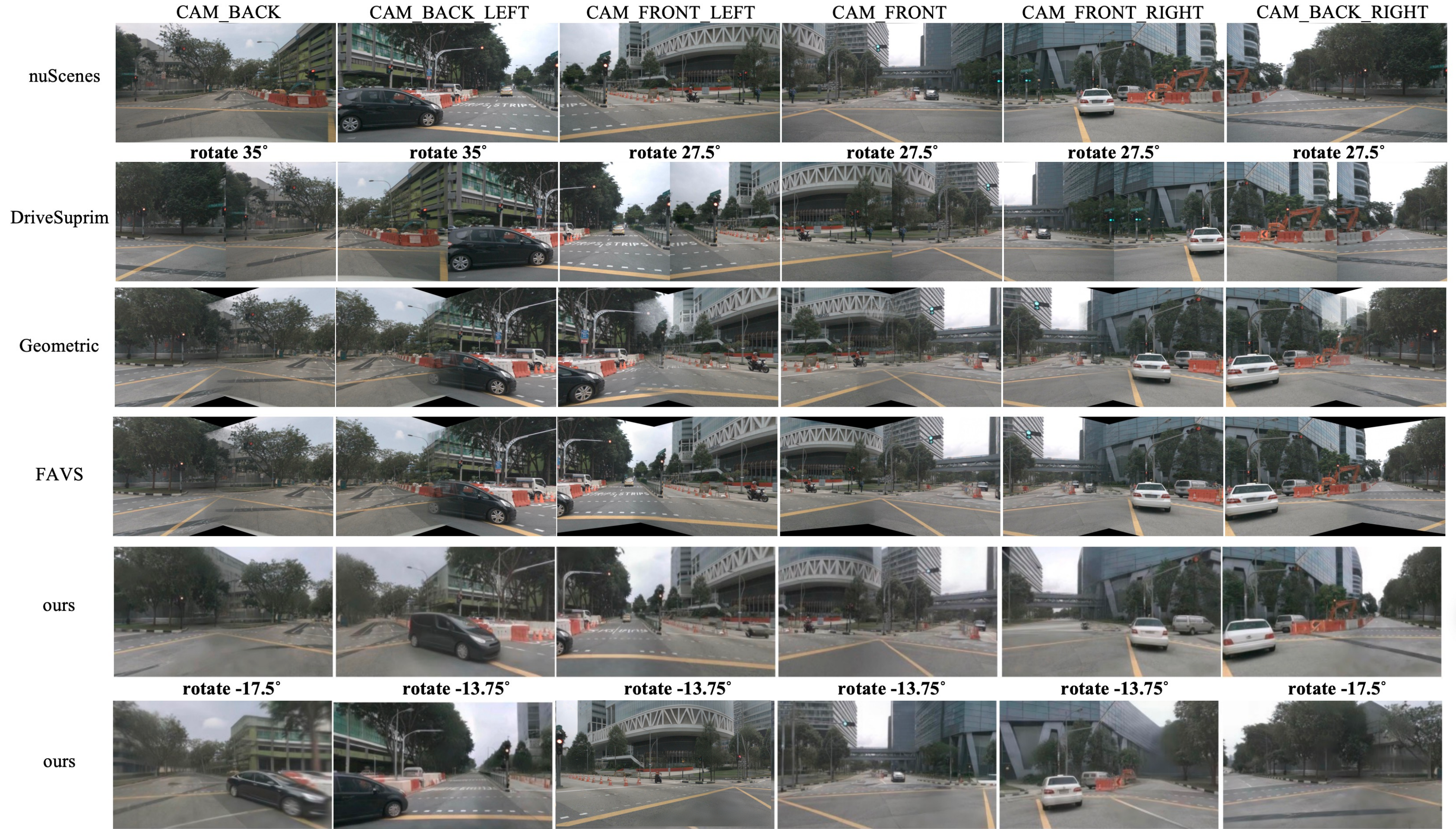}
\caption{\textbf{Qualitative comparison of novel-view synthesis under 27.5°, 35°, -13.75° and -17.5° camera rotations.} Row 1 shows the original 6-camera images from nuScenes~\cite{Caesar2020}. Row 2 displays results from DriveSuprim~\cite{Yao2025} using rotation-based augmentation. Row 3 displays synthesized views utilizing geometric transformations. Row 4 presents pseudo-views generated by our FAVS module. Row 5 illustrates the final results of our \arbiviewgen\, demonstrating improved consistency and realism in novel viewpoints.}
\label{fig:2}
\end{figure*}
\subsection{Dataset}

We conduct experiments on the nuScenes~\cite{Caesar2020} dataset, which contains 1,000 scenes, with each scene comprising approximately 40 keyframes sampled at 2Hz frequency, equipped with a complete sensor suite including 6 cameras, 5 radars, and 1 LiDAR. nuScenes~\cite{Caesar2020} provides high-precision LiDAR point cloud data and complete sensor calibration parameters, ensuring precise spatiotemporal alignment of multimodal data, making it an authoritative benchmark for autonomous driving scene understanding. We use 60\% of the scenes for training (approximately 24,000 frames), 20\% for validation, and 20\% for extrapolated viewpoint generation and colored point cloud evaluation, ultimately obtaining approximately 34,000 annotated frames. By coloring LiDAR point clouds using the original six cameras and projecting them to target viewpoints to generate sparse reference points, we construct a quantitative evaluation benchmark for extrapolated viewpoints.
\subsection{Metrics}
\begin{itemize} 
\item PSNR and SSIM measure reconstruction quality against LiDAR-projected reference points, with PSNR quantifying pixel-level fidelity and SSIM evaluating structural consistency.
\item MAE and RMSE assess pixel-wise accuracy, where MAE computes average absolute deviation and RMSE applies quadratic penalty for large errors. 
\end{itemize} 

\subsection{Novel View Evaluation}
\renewcommand{\arraystretch}{1.4}
\begin{table}[htbp]
\centering
\begin{tabular}{lcccc}
\hline 
\multirow{2}{*}{\textbf{Method}} & \textbf{Sparse-} & \textbf{Sparse-} & \textbf{Sparse-} & \textbf{Sparse-} \\
& \textbf{PSNR} & \textbf{SSIM} & \textbf{MAE} & \textbf{RMSE} \\
\hline 
DriveSuprim & 9.5647 & 0.8542 & 72.4672 & 87.5129 \\
\hline 
ArbiViewGen & \textbf{14.2335} & \textbf{0.9691} & \textbf{38.2820} & \textbf{49.5294} \\
\hline
\end{tabular}
\caption{\textbf{Quantitative comparison of novel-view image synthesis based on sparse ground-truth supervision.} Metrics are computed on sparse pixels projected from colored LiDAR point clouds. ArbiViewGen achieves significant improvements over the baseline DriveSuprim across all four metrics. }
\label{tab:results}
\end{table}
\noindent Since ArbiViewGen targets controllable arbitrary-view generation across diverse vehicle platforms, no prior method provides direct comparability. We adopt DriveSuprim, which applies rotation-based augmentation, as a reference baseline. Four sparse metrics (PSNR, SSIM, MAE, RMSE) are computed from colored LiDAR point clouds projected into novel views. ArbiViewGen consistently outperforms the baseline across all metrics, indicating superior fidelity and structural consistency under sparse supervision.

\renewcommand{\arraystretch}{1.4}
\begin{table}[htbp]
\centering
\begin{tabular}{lcccc}
\hline
\multirow{2}{*}{\textbf{Method}} & \textbf{Sparse-} & \textbf{Sparse-} & \textbf{Sparse-} & \textbf{Sparse-} \\
&\textbf{PSNR$\uparrow$} & \textbf{SSIM$\uparrow$} & \textbf{MAE$\downarrow$} & \textbf{RMSE$\downarrow$} \\
\hline
Geometric & 9.3167 & 0.8339 & 74.1309 & 88.7707 \\
FAVS & 11.8707 & 0.8813 & 42.4985 & 55.1446 \\
\hline
Ours & \textbf{14.2335} & \textbf{0.9691} & \textbf{38.2820} & \textbf{49.5294} \\
\hline
\end{tabular}
\caption{\textbf{Ablation study on key modules in ArbiViewGen.}“Geometric” refers to view projection without feature fusion; “FAVS” adds feature-aware stitching; “Ours” includes full cross-view consistency learning (CVC-SSL). Performance improves progressively with each component. }
\label{tab:ablation_study_2}
\end{table} 
\noindent We assess the contribution of each core component. As shown in Table~\ref{tab:ablation_study_2}, both the feature-aware stitching module (FAVS) and the cross-view consistency learning (CVC-SSL) yield clear performance gains. The full model achieves the best results across all metrics, confirming the effectiveness of our design.
\subsubsection{Visualization}
As shown in Figure~\ref{fig:2}, we visualize novel-view synthesis results under  27.5°, 35°, -13.75° and -17.5° camera rotations. DriveSuprim, trained with simple rotation-based augmentation, suffers from geometric distortions and object misalignment. The geometric projection baseline preserves rigid alignment but introduces tearing and black borders in regions where no original camera view provides information for the target viewpoint. FAVS improves alignment by leveraging camera poses and feature correspondences, yet still exhibits discontinuities and missing regions due to the lack of source-view content in those extrapolated directions. While not photorealistic, FAVS offers a coarse geometric prior that facilitates learning. With the full \arbiviewgen pipeline, the model generates more structurally consistent and spatially complete images, demonstrating better generalization to unseen viewpoints.

\section{Conclusion}
In this work, we introduce \arbiviewgen\ -- a controllable diffusion-based framework for arbitrary-view image generation in autonomous driving scenarios. By integrating a feature-aware stitching module (FAVS) and a cross-view consistency self-supervised learning strategy (CVC-SSL), our method effectively mitigates the challenge of lacking ground-truth supervision for extrapolated views, enabling arbitrary-view synthesis using only multi-camera images and pose information. The proposed framework enhances the adaptability and robustness of autonomous driving systems across various sensor configurations, facilitating cross-platform deployment and scalable data reuse. Despite promising experimental results, the framework still faces limitations in capturing fine-grained structural details in highly dynamic environments, particularly under sparse geometric constraints. Future work will focus on incorporating sparse-to-dense supervision signals, such as LiDAR-based depth priors and semantic consistency constraints, to further enhance the quality of novel-view generation.

\section{Acknowledgments}
This work was supported by the National Key R\&D Program of China, Project “Development of Large Model Technology and Scenario Library Construction for Autonomous Driving Data Closed-Loop” (Grant No.2024YFB2505501), and the Guangxi Key Scientific and Technological Project, Project “Research and Industrialization of High-Performance and Cost-Effective Urban Pilot Driving Technologies” (Grant No.Guangxi Science and Technology AA24206054)

\bibliography{aaai2026}

\clearpage
\appendix
\renewcommand{\thesection}{A}  % Section 编号变为 A
\renewcommand{\thesubsection}{A.\arabic{subsection}}
\section{Appendix} 
\addcontentsline{toc}{section}{A Appendix}
The appendix provides:  
1) a detailed explanation of the underlying latent diffusion mechanism employed in our method, including the formulation and role of each core component; and  
2) additional qualitative results comparing our generated multi-view images with baselines across a variety of camera viewpoints.

\subsection{Preliminaries}
\noindent
\subsubsection{Latent Diffusion Models}
Latent Diffusion Models (LDMs)~\cite{rombach2022high} serve as the foundation of our methodology. An LDM comprises three essential components: a variational autoencoder (VAE) ~\cite{kingma2013auto}with an encoder $\mathcal{E}$ and a decoder $\mathcal{D}$, a denoising network $\epsilon_\theta$, and a condition encoder $\tau_\theta$.

Given a high-resolution image $\mathbf{x} \in \mathbb{R}^{H \times W \times 3}$, the encoder $\mathcal{E}$ projects it into a lower-dimensional latent space, yielding $\mathbf{Z} = \mathcal{E}(\mathbf{x})$, where $\mathbf{Z} \in \mathbb{R}^{h \times w \times c}$. The down-sampling factor $f = H/h = W/w$ is typically set to 8 in widely used models such as Stable Diffusion (SD)~\cite{stablediffusion2022}. The latent representation can be mapped back to the image space by the decoder, i.e., $\tilde{\mathbf{x}} = \mathcal{D}(\mathbf{Z})$.

The training objective for LDMs is formulated as follows:
\begin{equation}
L_{\mathrm{LDM}} := \mathbb{E}_{\mathcal{E}(\mathbf{x}),\, \mathbf{y},\, \epsilon \sim \mathcal{N}(0,1),\, t} \left[ \left\| \epsilon - \epsilon_\theta(\mathbf{Z}_t,\, t,\, \tau_\theta(\mathbf{y})) \right\|_2^2 \right],
\end{equation}
where $t$ is uniformly sampled from $1$ to $T$, and $\mathbf{Z}_t$ denotes the noisy latent at time step $t$. The denoising network $\epsilon_\theta$ is a time-dependent U-Net~\cite{dhariwal2021diffusion} , enhanced with cross-attention mechanisms to incorporate the optional condition encoding $\tau_\theta(\mathbf{y})$. The condition $\mathbf{y}$ may represent a text prompt, an image, or any other user-specified input.

During inference, the denoising (reverse) process generates samples in the latent space, and the decoder reconstructs high-resolution images via a single forward pass. Furthermore, advanced samplers ~\cite{lu2022dpm,karras2022elucidating,song2020denoising} can be employed to accelerate the sampling process.

\subsubsection{The Multi-Branch U-Net}

To generate \( N \) different views, we uses \( N \) parallel U-Net branches~\cite{Tang2023}. These branches are not independent but are characterized by two key features:

\begin{itemize}
    \item \textbf{Weight Sharing}: All \( N \) U-Net branches share the exact same set of network weights. This means there is only a single copy of the U-Net parameters, which simultaneously processes \( N \) distinct inputs (the noisy latents for each of the \( N \) views). This design is highly parameter-efficient and crucially preserves the powerful generalization capabilities of the pre-trained Stable Diffusion model. 
    \item \textbf{Simultaneous Denoising}: The model takes the initial noisy latents for all views, \( \{\mathbf{Z}_t^{(1)}, \mathbf{Z}_t^{(2)}, \ldots, \mathbf{Z}_t^{(N)}\} \), and processes them through their respective branches concurrently. At each step of the reverse diffusion process, the model predicts the noise for all \( N \) views in parallel. This holistic approach fundamentally avoids the issue of error accumulation that is prevalent in autoregressive methods, where views are generated sequentially. 
\end{itemize}

\subsubsection{Correspondence-Aware Attention (CAA) for Consistency}

Parallel processing alone does not guarantee inter-view consistency. To ensure that objects and textures align seamlessly across different viewpoints, the method integrates the Correspondence-Aware Attention (CAA)~\cite{Tang2023} module. A CAA block is inserted after each U-Net block within the shared-weight architecture. It functions as a communication bridge between the parallel U-Net branches, forcing the model to consider cross-view relationships during generation.

The CAA mechanism operates as a targeted cross-view attention. For a given token at position \( \boldsymbol{s} \) in a source feature map \( \boldsymbol{F} \), the CAA block calculates attention scores by comparing it with corresponding tokens at positions \( \boldsymbol{t}' \) in a target feature map \( \boldsymbol{F}' \). This process is enhanced by incorporating positional encodings derived from the known geometric displacement between \( \boldsymbol{s} \) and its corresponding location in the target view, which explicitly informs the model about the spatial relationship. The resulting contextual information is then aggregated and fused back into the source feature, enriching it with multi-view context. A standard Feed-Forward Network (FFN)~\cite{vaswani2017attention}, a typical component of a transformer block, follows the attention layer to further process the integrated features.

By explicitly fusing information based on known camera poses, the CAA mechanism enforces consistency at every level of the U-Net. If view A and view B overlap, the CAA block ensures that the features generated for this overlapping region are coherent and aligned, leading to a consistent final multi-view output.
\subsubsection{Visualization Results}
\begin{figure*}[b]
\centering
\includegraphics[width=1.0\textwidth]{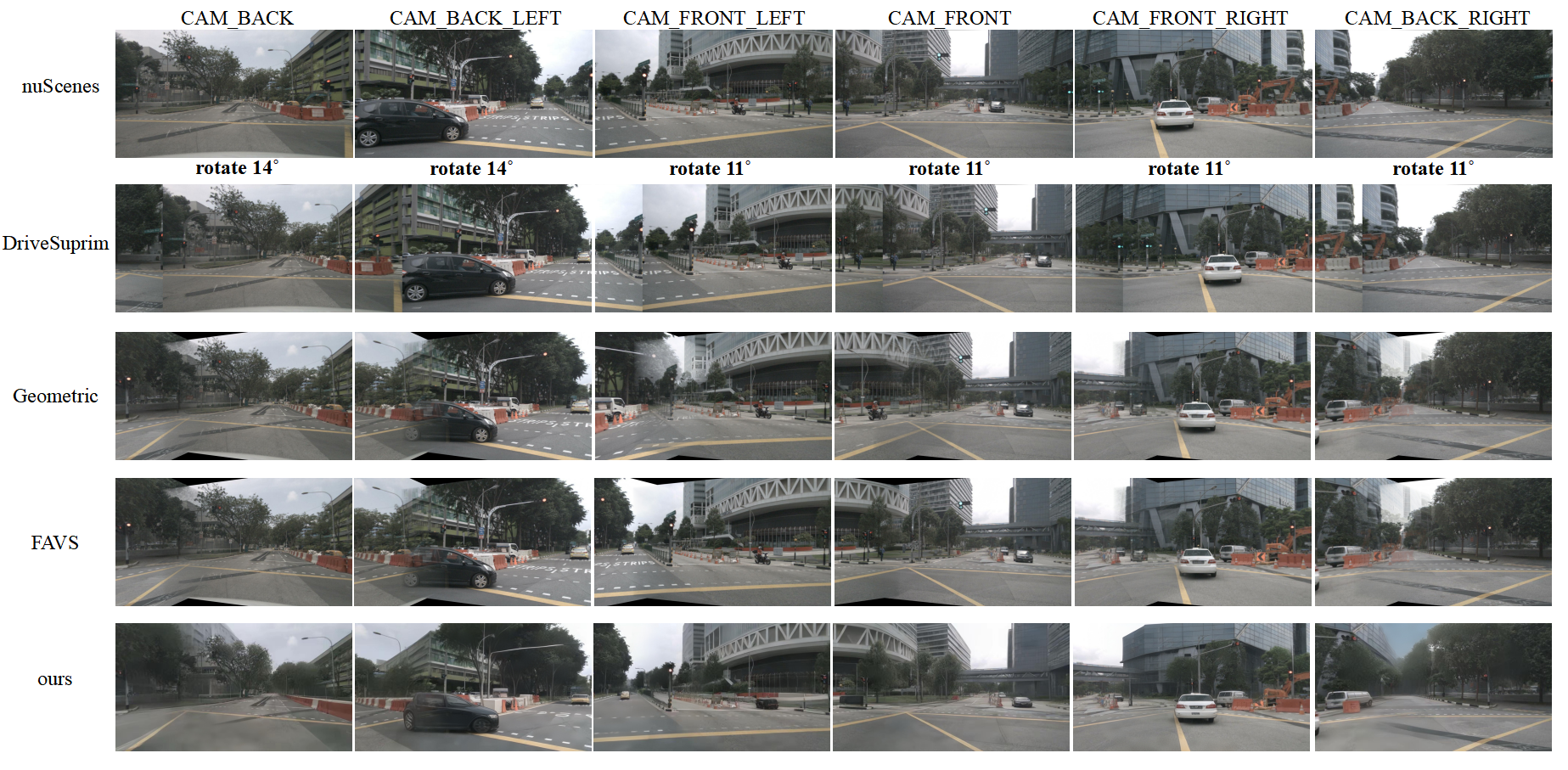}
\caption{\textbf{Qualitative comparison across six camera views.} Comparison shows that our method achieves better visual alignment and consistency than DriveSuprim, Geometric, and FAVS across different camera views (e.g., CAM\_BACK with rotate 14° and CAM\_FRONT with rotate 11°).}
\label{fig:3}
\end{figure*}

\begin{figure*}[t]
\centering
\includegraphics[width=1.0\textwidth]{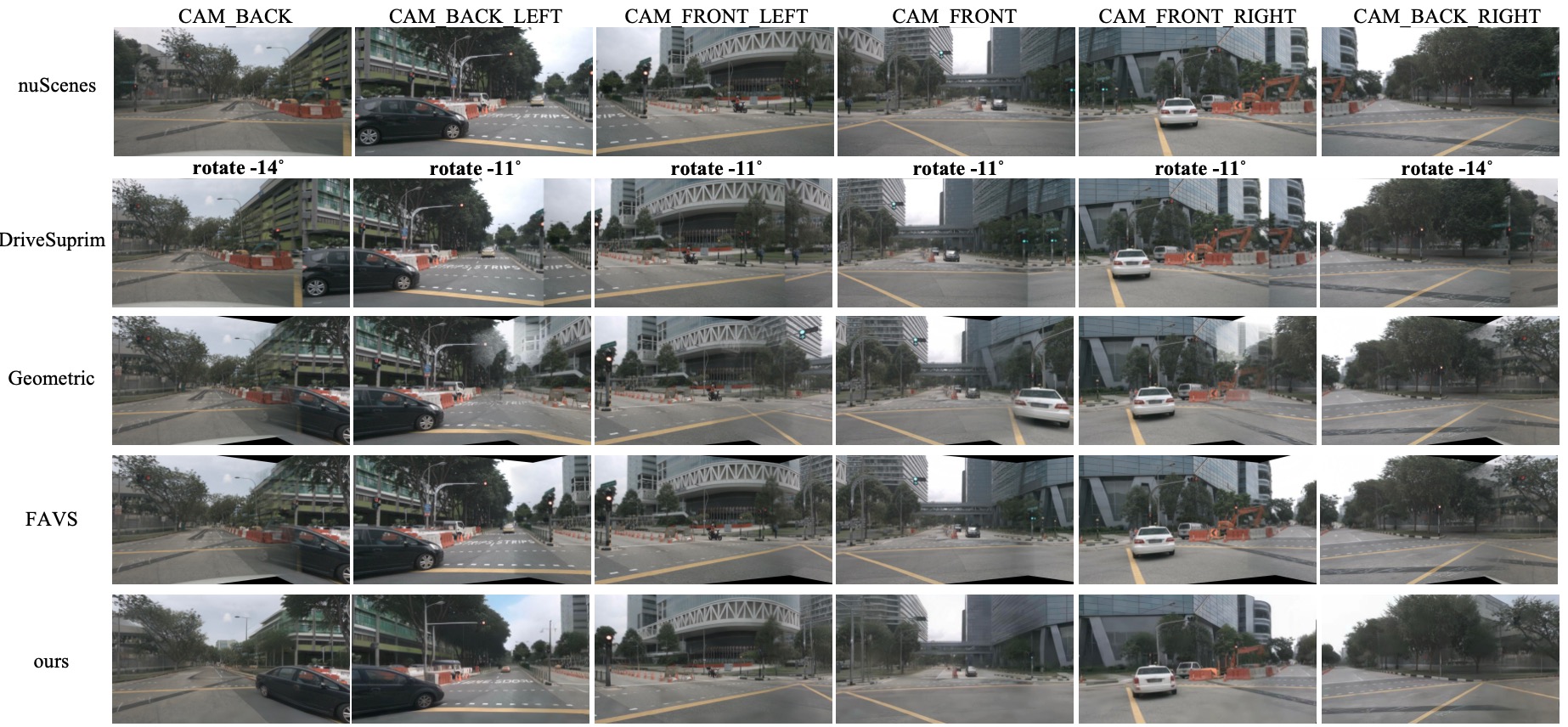}
\caption{\textbf{Qualitative comparison across six camera views.} Comparison shows that our method achieves better visual alignment and consistency than DriveSuprim, Geometric, and FAVS across different camera views (e.g., CAM\_BACK with rotate -14° and CAM\_FRONT with rotate -11°).}
\label{fig:4}
\end{figure*}

\begin{figure*}[t]
\centering
\includegraphics[width=1.0\textwidth]{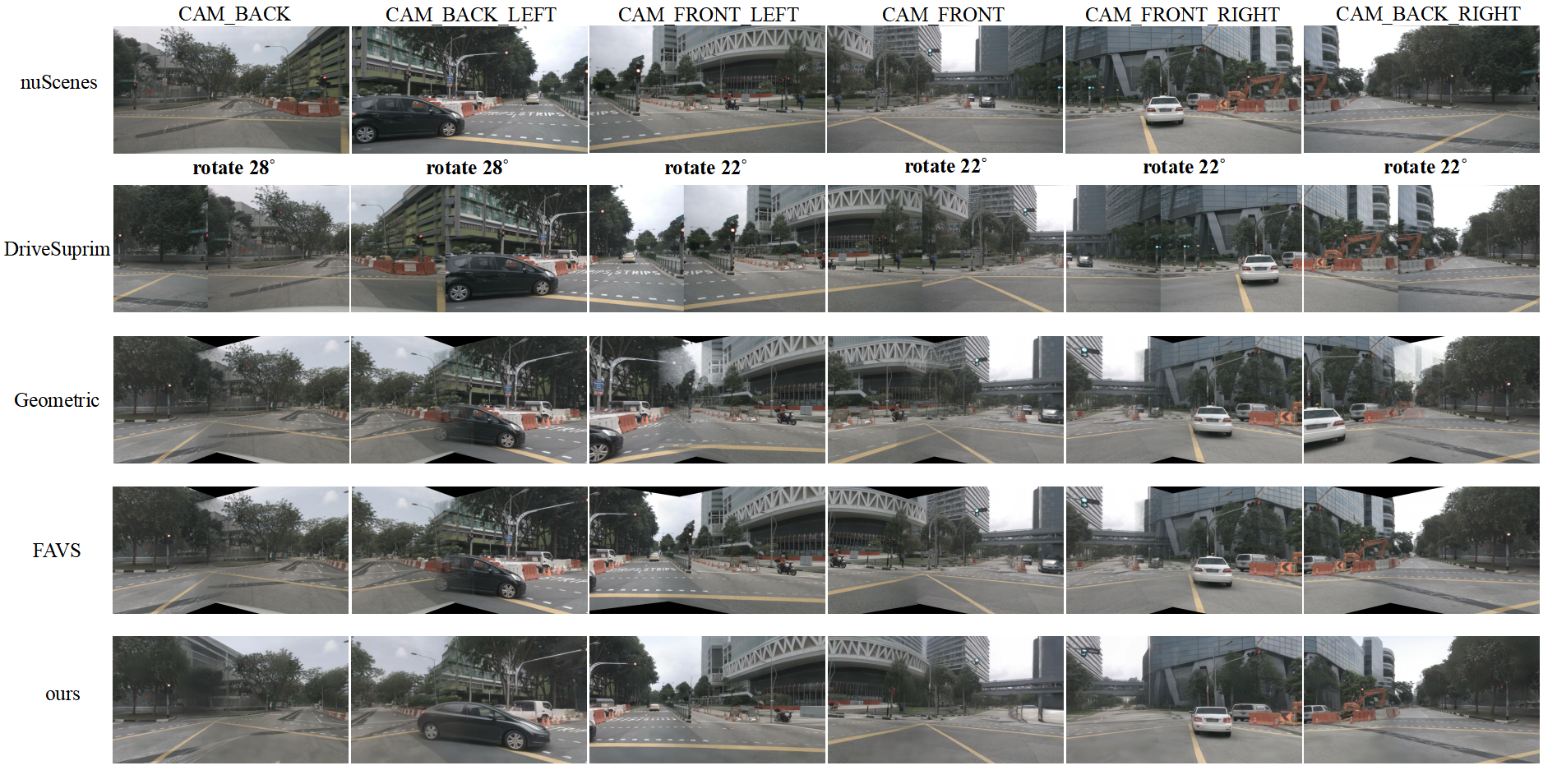}
\caption{\textbf{Qualitative comparison across six camera views.} Comparison shows that our method achieves better visual alignment and consistency than DriveSuprim, Geometric, and FAVS across different camera views (e.g., CAM\_BACK with rotate 28° and CAM\_FRONT with rotate 22°).}
\label{fig:5}
\end{figure*}

\begin{figure*}[t]
\centering
\includegraphics[width=1.0\textwidth]{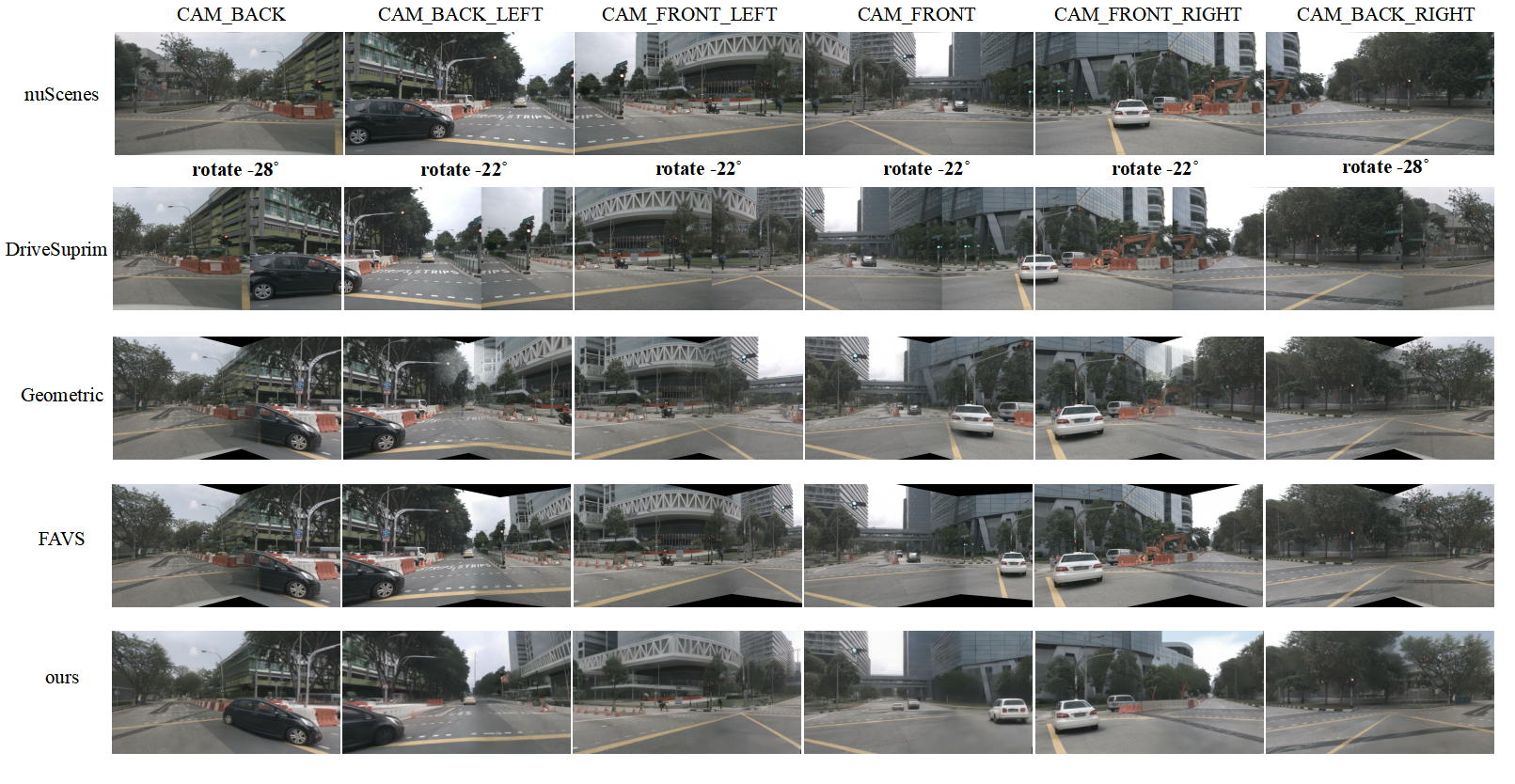}
\caption{\textbf{Qualitative comparison across six camera views.} Comparison shows that our method achieves better visual alignment and consistency than DriveSuprim, Geometric, and FAVS across different camera views (e.g., CAM\_BACK with rotate -28° and CAM\_FRONT with rotate -22°).}
\label{fig:6}
\end{figure*}

Figures~\ref{fig:3}--\ref{fig:6} provide additional qualitative results produced by ArbiViewGen. These results  demonstrate the effectiveness of our approach by generating plausible novel views in the absence of ground-truth images.
\clearpage

\end{document}